\documentclass[letterpaper, 10 pt, conference]{ieeeconf}  

\IEEEoverridecommandlockouts                              

\usepackage{graphics} 
\usepackage{graphicx}
\usepackage[svgnames]{xcolor}

\usepackage{amsmath} 
\usepackage{booktabs} 
\usepackage{balance} 
\usepackage{placeins} 
\usepackage{float}    
\usepackage{stfloats}
\usepackage{cuted}
\usepackage{amssymb}

\usepackage{subcaption}     
\newif\ifanonymous
\anonymousfalse

\title{\LARGE \bf
TransCurriculum: Multi-Dimensional Curriculum Learning for Fast \& Stable Locomotion
}

\ifanonymous
  \author{Anonymous}
\else
  \author{
  Prakhar Mishra$^{1,*}$,
  Amir Hossain Raj$^{2}$,
  Xuesu Xiao$^{2}$,
  and Dinesh Manocha$^{1}$%
  \thanks{$^{1}$University of Maryland, College Park, MD, USA.}%
  \thanks{$^{2}$George Mason University, Fairfax, VA, USA.}%
  \thanks{*Collaborating Researcher; M.Eng. University of Maryland, 2023.}%
  }
\fi

\begin{document}

\maketitle
\thispagestyle{empty}
\pagestyle{empty}

\begin{strip}
\vspace{-4em} 
\centering

\begin{minipage}{0.19\linewidth}
    \centering
    \includegraphics[width=\linewidth]{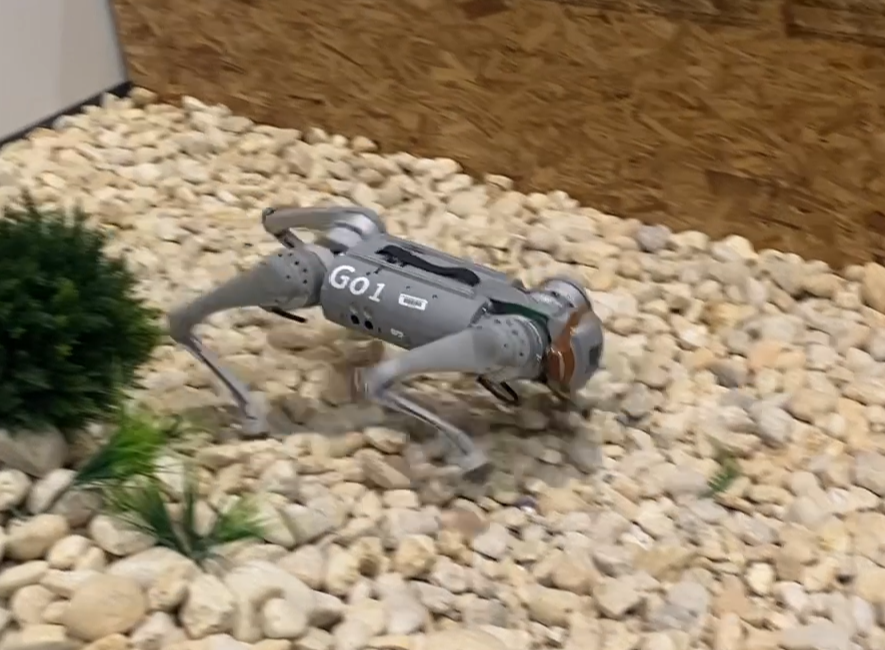}
\end{minipage}
\begin{minipage}{0.19\linewidth}
    \centering
    \includegraphics[width=\linewidth]{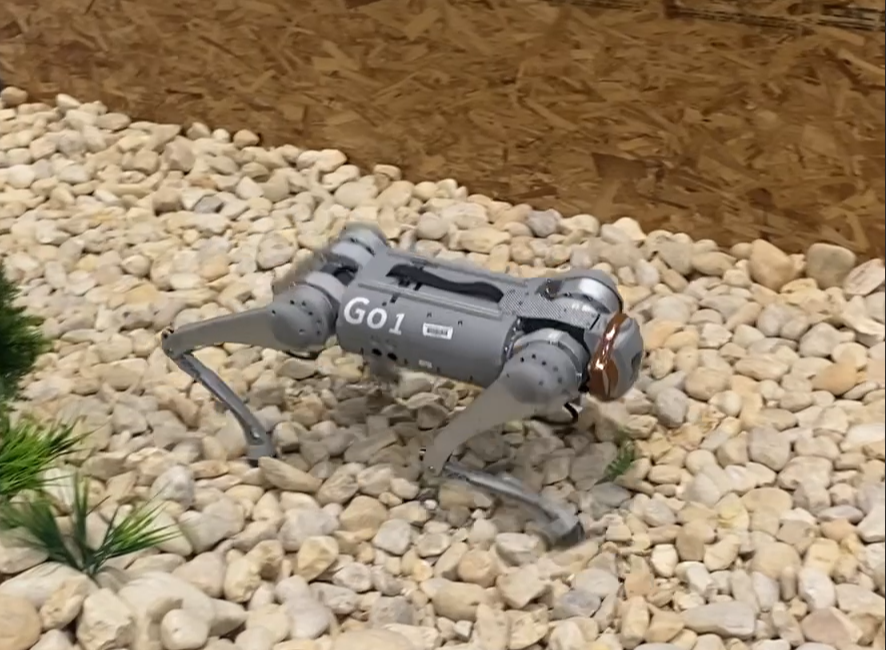}
\end{minipage}
\begin{minipage}{0.19\linewidth}
    \centering
    \includegraphics[width=\linewidth]{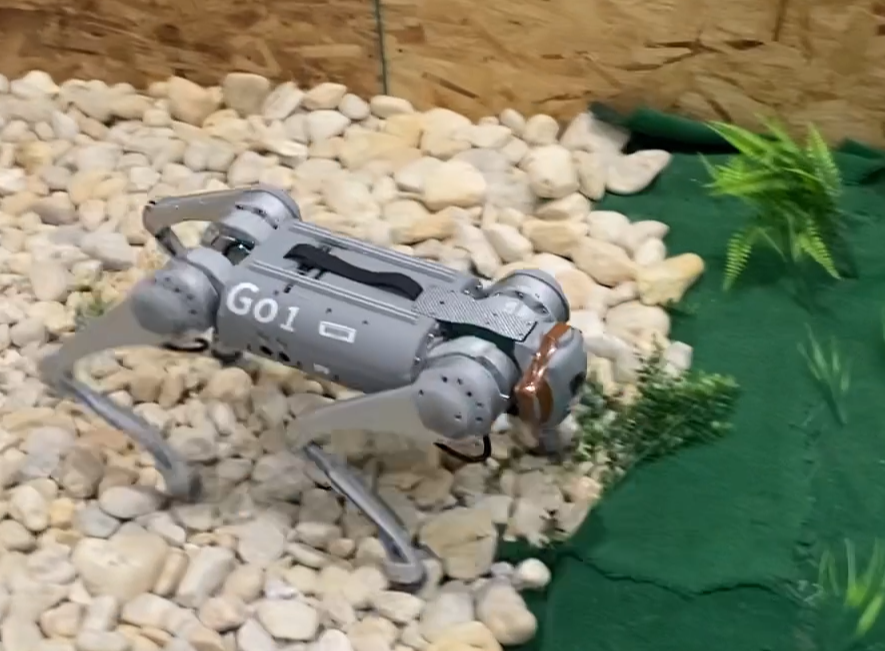}
\end{minipage}
\begin{minipage}{0.19\linewidth}
    \centering
    \includegraphics[width=\linewidth]{img/peb2.png}
\end{minipage}
\begin{minipage}{0.19\linewidth}
    \centering
    \includegraphics[width=\linewidth]{img/peb3.png}
\end{minipage}

\begin{minipage}{0.19\linewidth}
    \centering
    \includegraphics[width=\linewidth]{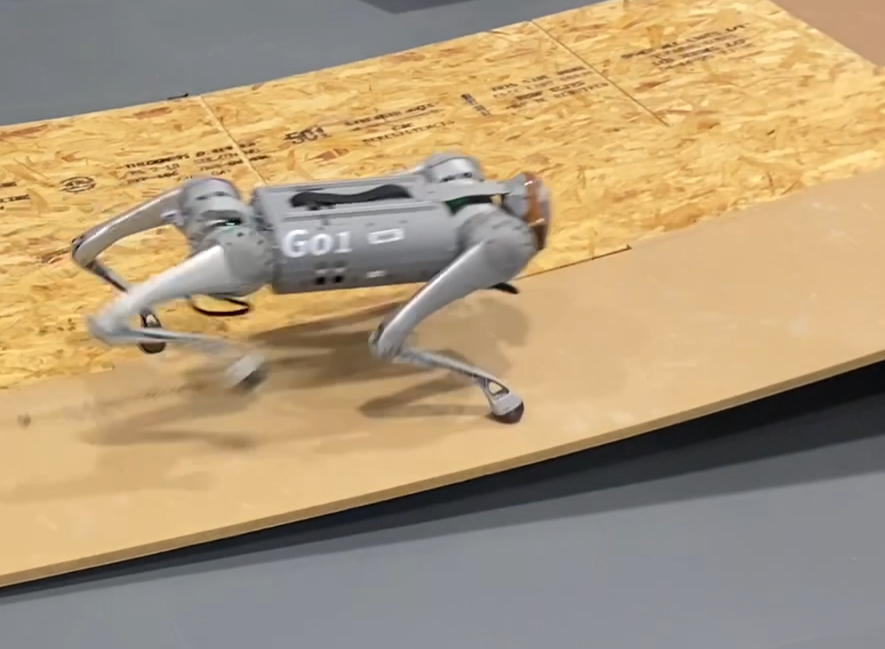}
\end{minipage}
\begin{minipage}{0.19\linewidth}
    \centering
    \includegraphics[width=\linewidth]{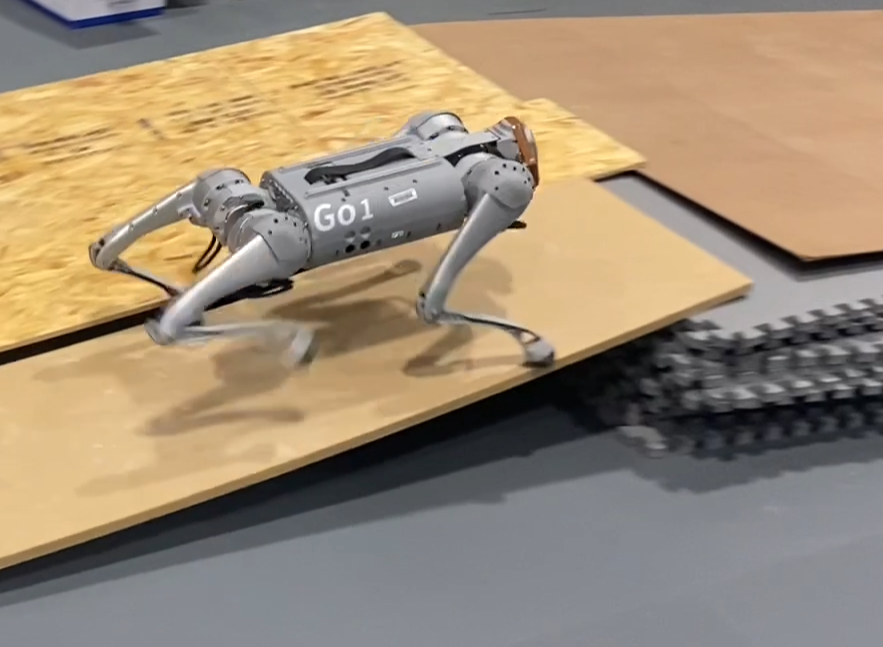}
\end{minipage}
\begin{minipage}{0.19\linewidth}
    \centering
    \includegraphics[width=\linewidth]{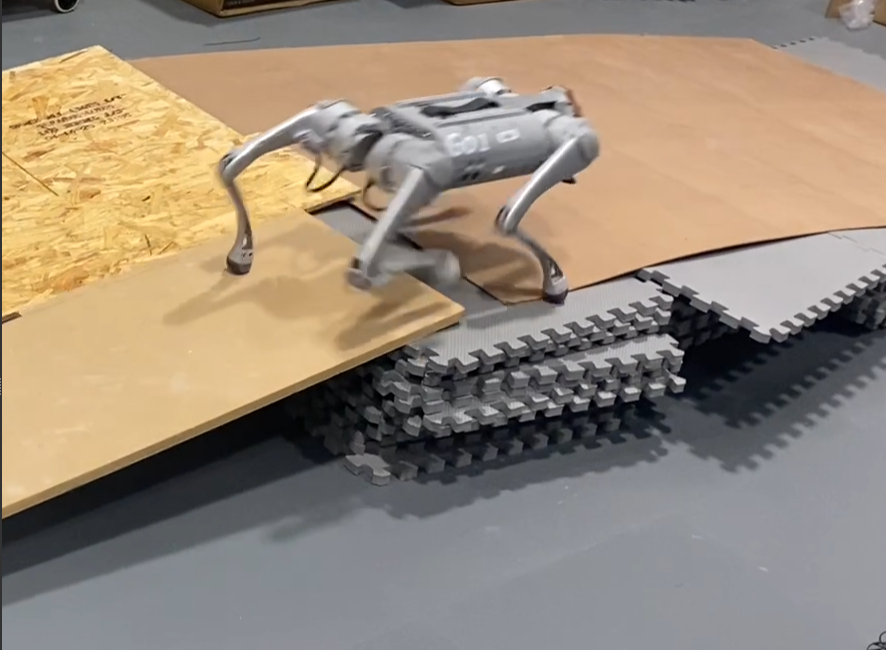}
\end{minipage}
\begin{minipage}{0.19\linewidth}
    \centering
    \includegraphics[width=\linewidth]{img/incl2.png}
\end{minipage}
\begin{minipage}{0.19\linewidth}
    \centering
    \includegraphics[width=\linewidth]{img/incl3.png}
\end{minipage}

\begin{minipage}{0.19\linewidth}
    \centering
    \includegraphics[width=\linewidth]{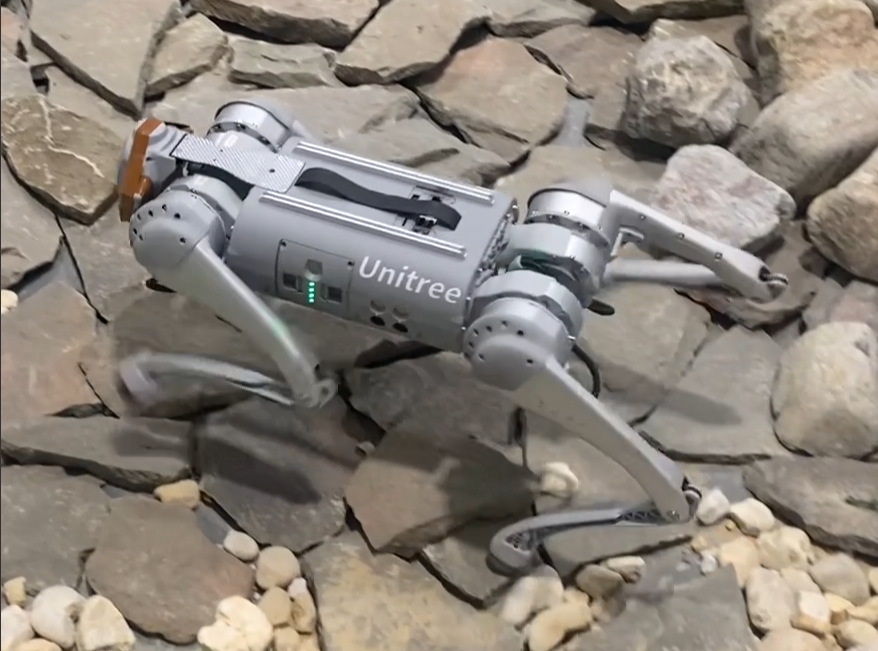}
\end{minipage}
\begin{minipage}{0.19\linewidth}
    \centering
    \includegraphics[width=\linewidth]{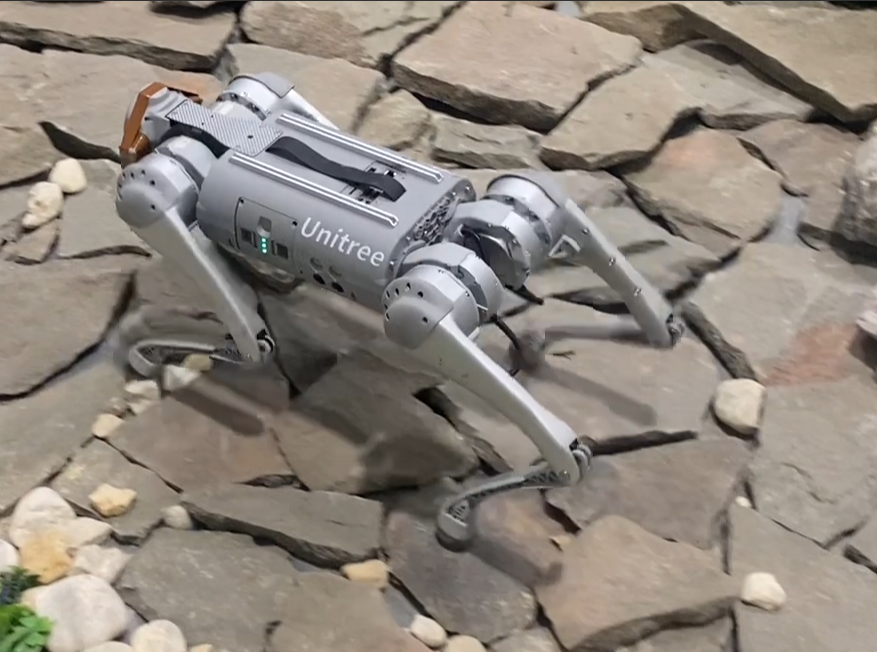}
\end{minipage}
\begin{minipage}{0.19\linewidth}
    \centering
    \includegraphics[width=\linewidth]{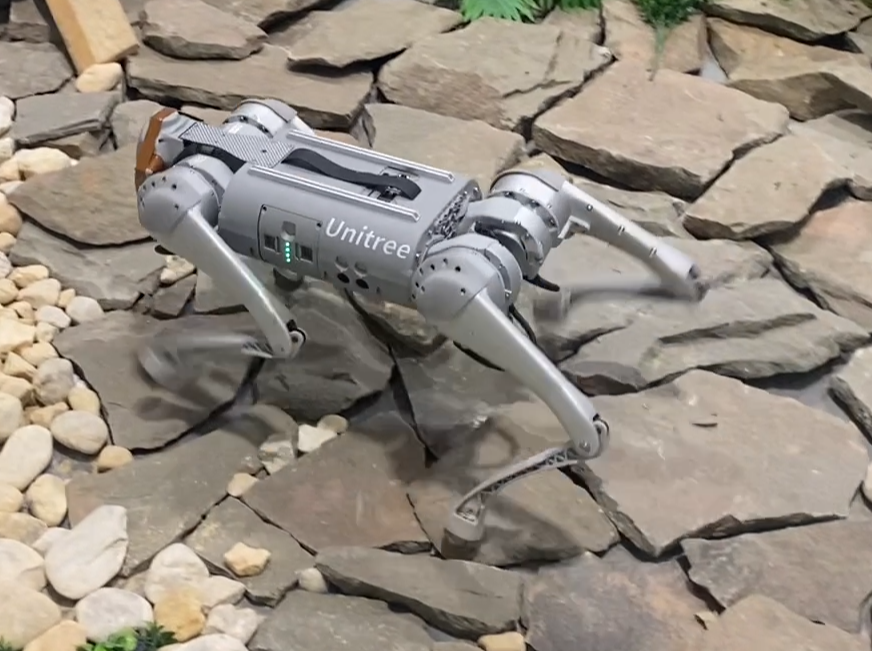}
\end{minipage}
\begin{minipage}{0.19\linewidth}
    \centering
    \includegraphics[width=\linewidth]{img/rocks3.png}
\end{minipage}
\begin{minipage}{0.19\linewidth}
    \centering
    \includegraphics[width=\linewidth]{img/rocks3.png}
\end{minipage}

\captionof{figure}{\textbf{Zero-shot hardware evaluation on diverse terrain (Unitree Go1, \emph{TransCurriculum}).}      
    We deploy Transcurriculum on Go1 and report speed, success rate and lateral deviation over short runs. \textbf{Row 1:} Pebbles (2-3 m) Go1 maintains $2.1 \pm 0.3$ m/s, 60\% success, $0.3 \pm 0.1$ m lateral deviation. 
    \textbf{Row 2:} Wooden slopes (approximate $20^\circ$ and 3-5 m) $3.1 \pm 0.4$ m/s, 80\% success, and lateral deviation of $0.5 \pm 0.3$ m.
    \textbf{Row 3:} Rocks (2-3 m) $1.5 \pm 0.4$ m/s, 50\%, success and $0.34 \pm 0.1$ m lateral deviation. The policy remains functional across terrain without finetuning, with some performance degradation with terrain difficulty,supporting benefits of history-aware \& multi-dimensional training.}
\label{fig:hardware_robust}
\end{strip}

\begin{abstract}

High-speed legged locomotion struggles with stability and transfer losses at higher command velocities during deployment. One of the key reasons is that most curricula vary difficulty along single axis, for example increase the range of command velocities, terrain difficulty, or domain parameters (e.g. friction or payload mass) using either fixed update rule or instantaneous rewards while ignoring how the history of how the robot training has evolved. We propose \emph{TransCurriculum}, a transformer based multi-dimensional curriculum learning approach for agile quadrupedal locomotion. TransCurriculum adapts to 3 key axes, namely velocity command targets, terrain difficulty, and domain randomization parameters (friction and payload mass). Rather than feeding task reward history information directly into the low-level control policy, our formulation exploits it at the curriculum level. A transformer-based teacher retrieves the sequence of observed rewards and uses it to predict the future rewards, success rate, and the learning progress to guide expansion of this multidimensional curriculum towards high performing task bins. And finally we validate our approach on the Unitree Go1 robot in a simulation (Isaac Gym) and deploy it zero-shot on the Go1 hardware. Our TransCurriculum policy achieves a maximum velocity of 6.3 m/s in the simulator and the highest stability score and outperforms prior curriculum baselines. We tested our TranCurriculum trained policy on various real world terrains (carpets, slopes, tiles, concrete), achieving a forward velocity of 4.1 m/s on carpet surpassing the fastest curriculum methods by nearly 18.8\% and achieves maximum zero-shot value among all tested methods (Table II). Our multi-dimensional curriculum also reduces the transfer loss to 18\% from 27\% for command only curriculum, demonstrating the benefits of joint training over velocity, terrain and domain randomization dimension while keeping the task success rate of 80--90\% on rigid indoor and outdoor surfaces.





\end{abstract}

\section{INTRODUCTION}

High speed legged locomotion in unstructured environments remains challenging because the performance is subjected to command velocity, terrain diversity, and unmodeled dynamic properties such as unknown friction, uneven terrain, slippery or inclined surfaces, etc. In the real world, legged robots operate without complete and accurate knowledge of these real world properties, which causes transfer loss, low task success rate, instability, or outright failure. Model-based controllers can model these environment parameters which can be effective in improving locomotion, but designing them requires substantial human expertise and may still fail to capture the full contact-rich dynamics \cite{zucker2010optimization}, \cite{hwangbo2019learning}, \cite{yin2007simbicon}. Recent reinforcement learning based methods have been shown to reduce this dependence on human expert in modeling these dynamics and have been shown to show strong robot performance on locomotion tasks such as running or walking, but training high speed RL policies remains a challenge as command velocity and environmental complexity increases \cite{margolis2024rapid}, \cite{rudin2022advanced}, \cite{xie2020allsteps}, \cite{hwangbo2019learning}, \cite{peng2020learning}. 


Curriculum learning has emerged as a bridge to improve high speed and stable locomotion task training \cite{bengio2009curriculum}. It has been used for an array of legged locomotion tasks such as command tracking, terrain difficulty progression, and other tasks such as gait type, climbing, etc \cite{margolis2024rapid}, \cite{margolis2023walk}, \cite{rudin2022learning}, \cite{rudin2022advanced}.
However, One of the key limitations of existing curricula is that they adapt along a single dominant axis either command velocity, terrain, or domain parameters utilizing fixed or threshold-based update rules \cite{margolis2024rapid}, \cite{zhang2024learning}, \cite{li2026scaling}. These methods work in narrow settings but often lose stability and fail to sustain rapid locomotion when the performance is highly dependent on multiple interacting factors such as commanded forward velocity, ground friction, mass, and terrain rather than completely isolated in real-world.


Another core difficulty with the multi-dimensional curriculum space is that scheduler cannot detect trends in policy learning and might over focus on already mastered task regions \cite{margolis2024rapid}, \cite{aractingi2023controlling}. So, this motivates us to use a history-aware curriculum to model temporal evolution of task rewards and adjust the curriculum accordingly rather than injecting directly into the low level controller.



\noindent {\bf Main Results:} 
To address this, we introduce \emph{\textbf{TransCurriculum}}, a \emph{\textbf{Trans}former-based \textbf{Curriculum} Learning} approach for fast and stable locomotion. TransCurriculum operates on a joint task space that includes commands, terrain, and domain-randomization parameters. Instead of directly injecting the reward history into the low-level actor-critic policy, we use it at the curriculum level. Specifically, we use a transformer-based teacher which retrieves training history, and predicts reward, success, and progress of the curriculum bins. And based on these predicted rewards, we influence sampling of the bins which improve learning. In this way, TransCurriculum reduces manual schedule design while improving the balance between agility, and stability via utilizing history.


\begin{figure}[thpb]
  \centering     
  \includegraphics[width=\columnwidth]{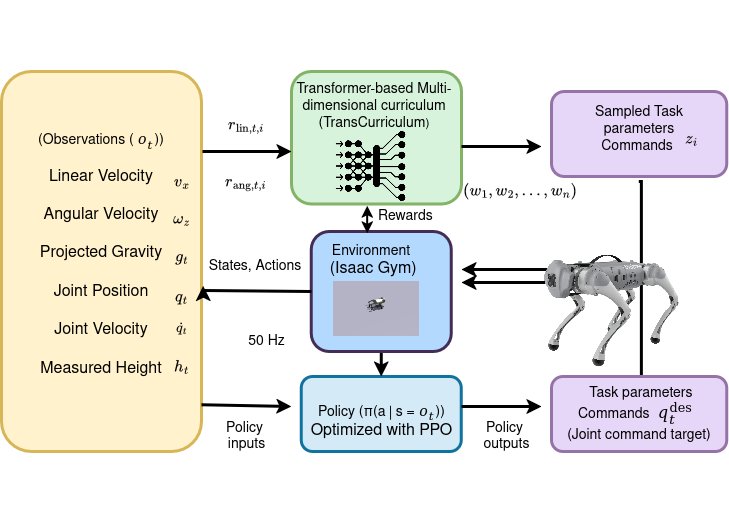}
  \caption{\textbf{TransCurriculum pipeline.} The low-level \textcolor{Indigo}{policy} $\pi_{\theta}$ is trained with \textbf{PPO} using rollouts from \textcolor{blue}{IsaacGym environment}. While the \textcolor{ForestGreen}{TransCurriculum module} maintains bins over multi-dimensional task space of commands, terrain difficulty and domain parameters and use rollouts ( \textcolor{Goldenrod}{observations and rewards}) to update distribution. The transformer-curriculum retrieves context-outcome history, to predict the rewards, success and progress of these curriculum bins. These sampled \textcolor{DeepPink}{task-context} $z$ is applied to the simulator for the next PPO rollout.}
  
 \label{fig:TransCurriculum_overview}
\end{figure}

\begin{itemize}
    \item \textbf{Method.} Unlike conventional single-axis curricula, TransCurriculum leverages locally retrieved training history to predict candidate-bin reward, success, and progress, over the velocity, terrain, and domain parameters enabling adaptive multidimensional curriculum expansion.
    
    \item \textbf{Approach.} Compared to fixed-rule or threshold based or single-axis curriculum \cite{margolis2024rapid}, \cite{li2026scaling}, \cite{aractingi2023controlling}, in TransCurriculum, we shift temporal modeling from the low-level control policy to the curriculum level and incorporate a multi-axis approach. And, this enables efficient bin sampling, higher speed, and lower sim-to-real transfer loss (ref. V. Results)
    
    \item \textbf{Evaluation.} We extensively evaluated TransCurriculum in simulation on the Unitree Go1 robot in an IsaacGym simulator and finally validated it on the Go1 hardware in real-world conditions across various terrains like rigid, deformable, irregular and moderate slopes.(ref. section IV-V)

    \item \textbf{Outcome.} In simulation, TransCurriculum achieves a velocity of $6.3 \pm 0.2$ m /s and $4.1 \pm 0.05$ m / s on Go1 robot. This is higher than the CHRL baseline ~\cite{li2024learning} of the command \& terrain curriculum ($3.45$ m / s) by $18.8\%$ and also the non-Go1 curriculum RLvRL ($3.9$ m / s) ~\cite{margolis2024rapid} with an improvement of $5.15\%$. TransCurriculum also improves stability  and reduces transfer loss from $27\%$ to $18\%$ relative to the command only curriuclum. Our results show that history-aware scheduling helps in improving the forward velocity while the multi-dimensional curriculum approach helps in stability and sim-to-real transfer loss (Section IV-V).
    
\end{itemize}

\section{Related Work}

\subsection{Curriculum Learning in Robotics} Curriculum learning has long been used for task stability and sample efficiency by controlling difficulty during training \cite{bengio2009curriculum}. Previous works include teacher-student curriculum, predefined curriculum, and the automatic curriculum generation for the parametrized environment \cite{matiisen2019teacher}, \cite{wang2021survey}. Curriculum learning has gained significant traction in robotics research, as these methods provide a useful foundation for robot learning, but they do not address the challenges of fast legged locomotion over command, terrain, and dynamics space \cite{luo2020accelerating}.

\subsection{Curriculum Learning for legged locomotion} 
Curriculum learning has gained some traction in legged locomotion, prior work has used it for a wide variety of curriculum strategies, namely command centric curricula for high speed tracking, terrain curriculum for rough terrain adaptation or hindsight\cite{aractingi2023controlling}, \cite{kumar2021rma}, \cite{chen2024reinforcement}, \cite{tidd2020guided}. RLvRL \cite{margolis2024rapid} uses fixed rules-based schedules, DreamWaq \cite{nahrendra2023dreamwaq} focuses on terrain aware locomotion, CHRL \cite{li2024learning} combines automatic curriculum with hindsight replay to learn locomotion control policy , but none exploit the reward history like TransCurriculum.


\subsection{Transfomer-based sequence modeling and curriculum design} 
Recent work has shown that transformers \cite{vaswani2017attention} can be incredibly helpful for legged locomotion, mainly for policy learning architectures \cite{radosavovic2024real}. The Terrain Transformer (TERT) \cite{lai2023sim} simply uses a transformer at the policy level for simulator-real transfer on diverse terrains. Body Transformer (BoT) \cite{sferrazza2024body} introduces embodiment-aware attention for policy learning. At the intersection of curriculum and transformers, CEC \cite{shi2023cross} injects ordered cross-episode experience and Decision transformer \cite{chen2021decision} models RL as return conditioned sequence modeling for curriculum like progression. None of these method model reward training history at multi-dimensional curriculum level for bin selection in locomotion, the core contribution of TransCurriculum.



\section{TransCurriculum: Multi-Dimensional Curriculum Learning}
Standard curriculum learning in legged robots separates tasks difficulty along single axis like command curriculum \cite{margolis2024rapid}, \cite{li2024learning}, terrain difficulty level \cite{rudin2022learning}, \cite{li2026scaling} or domain randomization \cite{li2024learning} parameters like friction, payload mass. Thereby quadrupedal robots struggle to sustain higher and stable velocities as the command-velocity range widens because in the real world these domains interact with each other. This decoupled approach gives a robot higher $v_{cmd}$ velocities with higher or medium friction, but fails when the friction is too low. Because the command curriculum never incorporated friction or payload observation along with progression in command velocities. We present \emph{TransCurriculum} which learns the multi-dimensional curriculum over the combined task space of commands, domain randomization, and terrain difficulty \eqref{eq:multidim}. \emph{TransCurriculum} acts as a curriculum-policy ($\pi_{\text{cur}}$), which selects curriculum bins to guide the low-level control policy ($\pi_{\theta}$).

\subsection{Tasks \& Design Space}

We define the training of the curriculum task \eqref{eq:multidim} as the intersection of command velocities $(c)$, domain parameters $(d)$ and terrain difficulty $(t)$. Where $(c)$ represents command velocity tasks $(v_{x}^{\text{cmd}}, v_{y}^{\text{cmd}}, \omega_{z}^{\text{cmd}})$, $d$ represents domain randomization features such as friction $(\mu)$ and payload mass $(m)$ and $t$ is terrain difficulty $t \in [0,1]$.

\begin{equation}
z = [c, d, t] \in \mathbb{R}^{D}
\label{eq:multidim}
\end{equation}

This multidimensional representation of the curriculum allows the policy to reason over multiple sources of difficulties within the defined task space before proceeding to the higher task difficulty level. And the curriculum design space is discretized into a grid of bins centroids $G = \{ g_i \}_{i=1}^{M}$, where $M$ is the product of bin count per task dimension. In our experiments, we divided the space into $20 \times 10 \times 20 = 4000 \text{ bins}$ (Section IV and fig.4 justify this granularity). 

\subsection{Curriculum Distribution and Bin Sampling}

Each bin $i$ has weight of $w_i$ demonstrating the current emphasis of the curriculum on that bin, and the sampling probability distribution of the bin $i$ is given by \eqref{eq:Curriculum sampling policy}:

\begin{equation}
\pi_{\mathrm{cur}}(i) = \frac{w_i}{\sum_{j=1}^{M} w_j}
\label{eq:Curriculum sampling policy}
\end{equation}

Once a bin $i$ is selected, we draw the specific task context $z_i$ uniformly:
\begin{equation}
i \sim \pi_{\mathrm{cur}}, \qquad z \sim \mathrm{Uniform}(\mathrm{cell}(g_i))
\label{eq:uniformCurriculum}
\end{equation}

And once the context space $z_i$ is sampled, it is decomposed into the respective command, DR parameters, and terrain difficulty and is instantiated in the simulation environment. As training progresses, the weights $w_i$ of the high performing bins increase based on the predicted rewards of the transformer. In addition, neighboring bins are also assigned higher weights according to their expected rewards, facilitating exploration.

\subsection{Episode Outcomes and EMA based progress tracking }
At the end of each episode, we have an outcome vector summarizing policy's performance:

\begin{equation}
y = [r_{\mathrm{lin}},\, r_{\mathrm{ang}},\, f,\, \ell]
\label{eq:outcome}
\end{equation}

where $r_{lin}$ \&  $r_{ang}$ are linear and angular tracking rewards, $f$ is binary fall indicator, and $\ell$ is the duration of normalized episodes. We derive a binary success indicator label $s$ that demonstrates whether the policy met the minimum tracking thresholds:

\begin{equation}
s = \mathbb{1}\!\left[
r_{\mathrm{lin}} > \tau_{\mathrm{lin}}
\;\land\;
r_{\mathrm{ang}} > \tau_{\mathrm{ang}}
\right]
\label{eq:success}
\end{equation}

We also compute a \emph{progress signal} to capture whether $\pi_{{\theta}}$ is improving in that region defined as the difference between the current average reward and an exponential moving average (EMA) maintained for each bin, where $\alpha \in [0,1]$ is the EMA smoothing rate.

\begin{equation}
r = \frac{r_{\mathrm{lin}} + r_{\mathrm{ang}}}{2},
\qquad
p = r - \bar{r}_i,
\qquad
\bar{r}_i \leftarrow \alpha r + (1-\alpha)\bar{r}_i
\label{eq:progress}
\end{equation}

This progress signal is essential for differentiating local improvement from episode noise, combined with the Transformer's ability to generalize across neighboring bins, enabling the curriculum to focus on most productive bins.

\subsection{History retrieval }

We maintain a memory buffer of all past task-outcome pairs stored during training:

\begin{equation}
\mathcal{H} = \{(z_j, y_j)\}_{j=1}^{N}
\label{eq:history}
\end{equation}

When evaluating the bin with context $z$, we retrieve its $k$ nearest neighbors from this history. 

\begin{equation}
H(z) = \mathrm{KNN}_k(\mathcal{H}, z)
\label{eq:knn}
\end{equation}

And this local retrieval mechanism forces the teacher to learn context-dependent relationships such as how rewards change with terrain or dynamics for the given command values rather than relying on coarse global training values (such as a single EMA over all bins). As a result, this forces the teacher to reason locally and generalize from well-explored regions to their unexplored neighbors.

\subsection{Transformer Teacher }

Given a candidate context $z$ and its retrieved local history $H(z)$, the transformer teacher predicts three quantities: expected reward or predicted reward $\hat{r}(z)$, success probability $\hat{s}(z)$, and progress $\hat{p}(z)$:

\begin{equation}
\big(\hat{r}(z),\, \hat{s}(z),\, \hat{p}(z)\big)
=
f_{\psi}\!\big(H(z), z\big)
\label{eq:teacher_prediction}
\end{equation}

The history tokens encode the $k$ retrieved context-outcome pairs ($z_j$,$y_j$), capturing the temporal evolution of the training performance of the neighboring bins. The query token attends to the retrieved tokens via cross-attention, enabling the teacher to infer whether the given bin is likely to be useful, already mastered, or still improving.

\textbf{Why Transformer?} A feedforward network (MLP) received all the tracking rewards input and predict bin features and cannot model how the rewards per bin evolve over time. A recurrent neural network (for e.g. RNN) can model this via hidden layer but might still miss the most effective past outcomes. A transformer's cross attention helps in learning those hidden pattern across different time frame. Our ablation (Table III) also confirms that history-aware architectures (Transformer, RNN) dramatically outperform non-history methods like MLP, while Transformer outperforms in terms of speed, stability and task success rate. 

\subsection{Reward-based curriculum expansion }

TransCurriculum expands the active curriculum outward from bins where the policy has already demonstrated empirical success. This ensures that our expansion strategy is grounded in the policy rather than solely relying on transformer predictions. We select the bins whose tracking rewards exceed a defined threshold:

\begin{equation}
S_t =
\left\{
i :
r_{\mathrm{lin},i} > \tau_{\mathrm{lin}}
\;\land\;
r_{\mathrm{ang},i} > \tau_{\mathrm{ang}}
\right\}
\label{eq:successful_bins}
\end{equation}

Next, we add the local neighboring bins ($\mathcal{N}(.)$) of the successful bins:

\begin{equation}
C_t = S_t \cup \mathcal{N}(S_t)
\label{eq:candidate_set}
\end{equation}

For curriculum expansion, we used larger neighborhood radii along velocity command dimensions for faster expansion toward higher command range, compared to smaller range along terrain and domain parameters for a gradual robustness growth. For each bin $i$, our transformer predicts the expected reward:

\begin{equation}
\hat{r}_i =
\begin{cases}
\dfrac{\hat{r}_{\mathrm{lin},i} + \hat{r}_{\mathrm{ang},i}}{2},
& \text{if } |\mathcal{H}| \ge k, \\[8pt]
1.0,
& \text{otherwise}
\end{cases}
\label{eq:pred_reward_score}
\end{equation}

The fallback value of 1.0 helps the curriculum recover in an early stage when the history buffer is too small to make any reliable local retrieval. But as training progress and the history buffer grows, the teacher's prediction increasingly guide the curriculum bin expansion. 

\begin{equation}
w_i \leftarrow
\mathrm{clip}
\!\left(
w_i + \beta_0 + \beta_1 \max(\hat{r}_i, 0),
\, 0,\, 1
\right),
\qquad i \in C_t
\label{eq:weight_update}
\end{equation}

The predicted rewards of the transformer $\hat{r}_i$ decide the direction of curriculum expansion (higher predicted rewards lead to a higher weight increase for those bins, along with their neighboring bins). This design prevents curriculum from overfitting to local maxima or ignore underexplored regions.

\begin{table}[t]
\centering
\caption{TransCurriculum Algorithm}
\label{tab:transcurriculum_algo}
\begin{tabular}{p{0.95\linewidth}}
\hline
\textbf{Input:} curriculum weights $w_i$, history buffer $\mathcal{H} \gets \emptyset$, transformer $f_{\psi}$ \\[2pt]
\hline
\textbf{for} each PPO iteration \textbf{do} \\
\quad \textbf{for} each parallel environment $k$ \textbf{do} \\
\qquad \textbf{if} episode ended \textbf{then} \\[3pt]
\qquad\quad \textit{ \textbf{Curriculum Update Rule}} \\
\qquad\quad 1. Successful bins: $\mathcal{S}_t \gets \{r_{\text{lin},i} > \tau_{\text{lin}} \wedge r_{\text{ang},i} > \tau_{\text{ang}}\}$ \hfill [Eq.~10] \\
\qquad\quad 2. Build neighbors: $\mathcal{C}_t \gets \mathcal{S}_t \cup \mathcal{N}(\mathcal{S}_t)$ \hfill [Eq.~11] \\
\qquad\quad 3. \textbf{for} $i \in \mathcal{C}_t$: predict $\hat{r}_i \gets f_{\psi}(\mathcal{H}(z_i), z_i)$ \hfill [Eq.~8, 9, 12] \\
\qquad\quad 4. Update: $w_i \gets \text{clip}(w_i + \beta_0 + \beta_1 \max(\hat{r}_i, 0),\; 0, 1)$ \hfill [Eq.~13] \\
\qquad\quad 5. Update EMA, success; append outcomes to $\mathcal{H}$ \hfill [Eq.~5--6] \\
\qquad\quad 6. Train $f_{\psi}$ on current batch \hfill [Eq.~14] \\[3pt]
\qquad\quad \textit{\textbf{Task Sampling}} \\
\qquad\quad 7. Draw bin $i \sim \pi_{\text{cur}}$, sample $z \sim \text{Uniform}(\text{cell}(g_i))$ \hfill [Eq.~2--3] \\
\qquad\quad 8. Instantiate $z$: set commands, friction, mass, terrain \\[2pt]
\qquad \textbf{end if} \\
\qquad Execute $\pi_{\theta}(a_t \mid o_t)$ \\
\quad \textbf{end for} \\
\quad PPO update on $\pi_{\theta}$ \quad \textit{(no gradients to $f_{\psi}$)} \\
\textbf{end for} \\
\hline
\end{tabular}
\end{table}

\subsection{Teacher optimization }

We train our transformer teacher using a multi-task loss function that supervises 3 prediction heads \eqref{eq:teacher_loss} 

\begin{equation}
\mathcal{L}_{\mathrm{teacher}}
=
\underbrace{\left\| \hat{r} - r \right\|_2^2}_{\mathcal{L}_{\mathrm{reward}}}
+
\underbrace{\mathrm{BCEWithLogits}(\hat{s}, s)}_{\mathcal{L}_{\mathrm{success}}}
+
\lambda
\underbrace{\left\| \hat{p} - p \right\|_2^2}_{\mathcal{L}_{\mathrm{progress}}}
\label{eq:teacher_loss}
\end{equation}

The \emph{reward head} (MSE Loss) aims to minimize the error between the observed rewards $(r_t)$ and the predicted rewards $\hat{r}(z)$. The \emph{success head} (binary cross-entropy) tracks whether the bins meet the minimum tracking threshold or not. The \emph{progress head} (weighted MSE) predicts whether the performance of a particular bin is stagnant, improving, or declining. As the training history grows, the teacher's prediction becomes increasingly accurate and the curriculum expansion becomes more targeted, since we use those signals to update the curriculum weight $(w_i)$. This allows for more efficient bin selection and faster convergence to higher and more stable forward velocities.

\begin{table*}[t]
\centering
\caption{TransCurriculum vs. representative curriculum baselines. Zero-shot transfer robot and maximum zero-shot real-world speed are reported when available (\textsuperscript{†}).}
\label{tab:method_comparison}
\renewcommand{\arraystretch}{1.1}
\setlength{\tabcolsep}{5pt}
\begin{tabular}{lcccccc}
\toprule
Method &
Curriculum target &
Curriculum signal &
Curriculum History modeling &
\begin{tabular}[c]{@{}c@{}}Zero-shot\textsuperscript{†}\\robot\end{tabular} &
\begin{tabular}[c]{@{}c@{}}Max zero-shot\textsuperscript{†}\\speed (m/s)\end{tabular} \\
\midrule

RMA~\cite{kumar2021rma} &
None &
None &
None &
A1 &
--- \\

RLvRL~\cite{margolis2024rapid} &
Velocity commands &
Fixed-rule &
EMA  &
Mini Cheetah &
3.9 \\

Risky Terrains~\cite{zhang2024learning} &
Terrain difficulty &
Task success &
None  &
ANYmal &
$ 2.5$ \\

DreamWaQ~\cite{nahrendra2023dreamwaq} &
Terrain difficulty &
Curriculum heuristics &
None &
A1 &
3.0 \\

CHRL~\cite{li2024learning} &
Commands + Terrain &
Hindsight success &
Replay &
Custom &
3.45 \\

LP-ACRL~\cite{li2026scaling} &
Commands + Terrain  &
Learning progress &
Online estimator &
ANYmal &
2.5 \\

Aractingi~\cite{aractingi2023controlling} &
Reward + Terrain &
Obs--Reward&
None &
Solo12 &
1.5 \\

\textbf{TransCurriculum (Ours)} &
Commands + Terrain + DR &
Pred--Reward &
Transformer &
Go1 &
\textbf{4.1} \\

\bottomrule
\end{tabular}

\vspace{2pt}
\footnotesize
\textsuperscript{†}\,Reported zero-shot transfer robot and maximum zero-shot real-world speed from the corresponding cited paper, when available.\\
\emph{Legend:}
Obs--Reward = observed reward ;
Pred--Reward = predicted reward from the curriculum model.
\vspace{-1em} 
\end{table*}





\section{Experiments}

\subsection{Simulation Environment}
\textbf{Simulation Details.} We implement TransCurriculum on top of open-source repositories \cite{margolis2024rapid} and \cite{rudin2022learning}, using the IsaacGym simulator \cite{makoviychuk2021isaac}. Our primary platform is Unitree Go1 (12 DoF), modeled from the manufacturer's URDF. 

\textbf{Training budget.} All our training experiments use $4000$ parallel environments at a control frequency of 200 Hz (\(\mathbf{\Delta t}=5\;{\rm ms}\)) on a single Nvidia RTX 4090 laptop-based GPU. Training runs 400 million steps ( roughly 4000 PPO updates) in about 4 hours.

\textbf{Command Curriculum.} Following \cite{margolis2024rapid}, we also initialize command velocities from the lower range $[-1.0,1.0]$ and gradually expand by $\pm 0.5$ as the policy starts to improve. Starting with a wide initial range, such as $[-5,5]$, destabilizes learning \cite{margolis2023walk}, \cite{kumar2021rma}, where as narrow range earlier tend to produce a more stable and faster speed (Fig. 3).


\textbf{Discretization.} The joint task space is normalized to $[-1, 1]$ and divided into $20 \times 10 \times 20$ bins, generating a total of $M = 4000 \text{ bins}$. We ablate this choice across 250, 1000, 4000 and 6000 bins and found that 4000 is the sweet spot for sample efficiency and speed optimization (Figure 4).


\textbf{Domain Randomization (DR).} To facilitate sim-to-real transfer, we randomize key dynamics parameters (e.g. friction and payload mass)\cite{tobin2017domain}. Rather than widening these ranges, which can produce conservative policy behavior \cite{margolis2024rapid}, \cite{rudin2022learning}, TransCurriculum included these DR ranges along the curriculum axis \eqref{eq:multidim}. Thereby controlling when the policy is exposed to harder dynamics parameters based on demonstrated competence.



\begin{figure}[thpb]
  \centering
  \includegraphics[width=0.9\columnwidth]{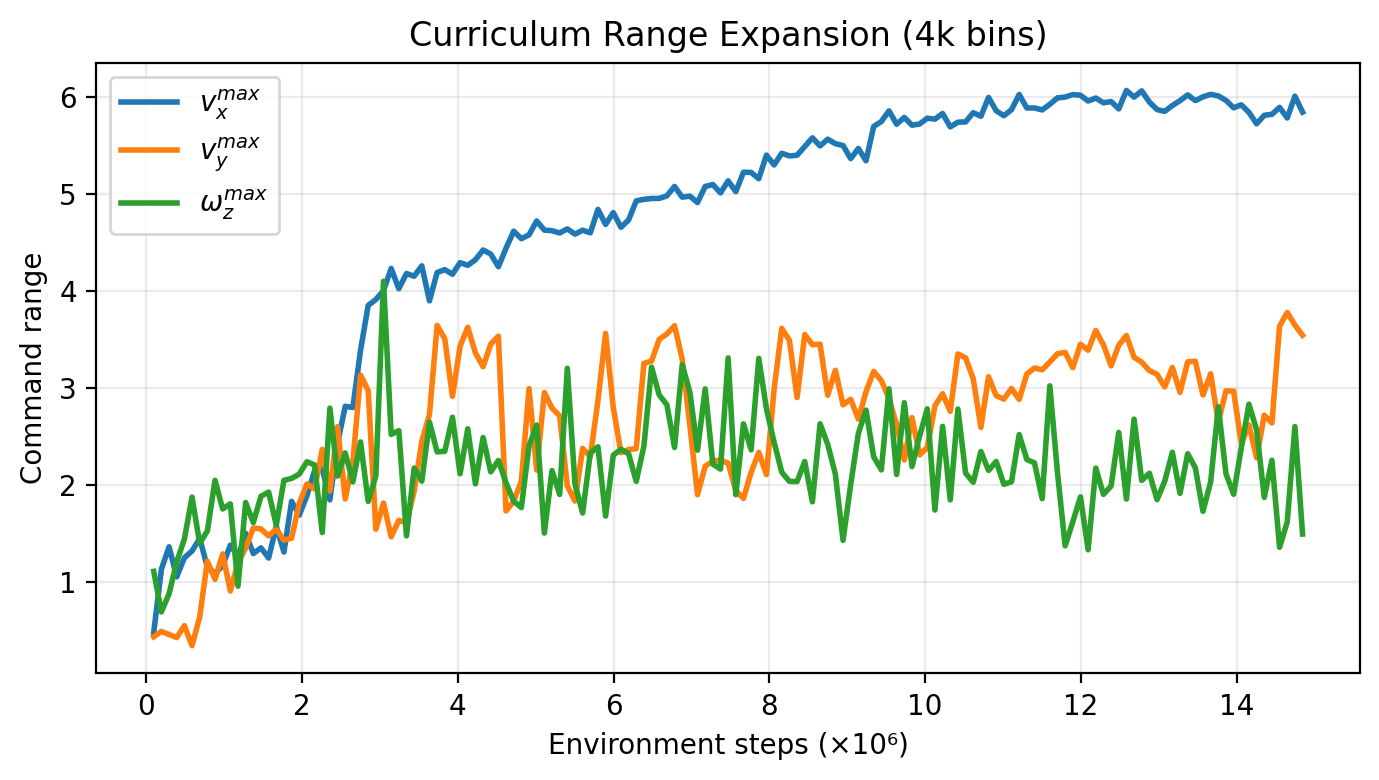}
  \caption{\textbf{Curriculum range expansion over command space:} 
  TransCurriculum starts from a narrow command range of $[-1.0,1.0]$ and gradually expands by  $[-0.5, 0.5]$ as the policy achieves stable velocity tracking performance. We plot the maximum sampled command values during training, \textcolor{blue}{$v_x^{\text{max}}$}, \textcolor{orange}{$v_y^{\text{max}}$} and \textcolor{green}{$\omega_z^{\text{max}}$}, as TransCurriculum expands the command range. The curriculum steadily expands along forward speed $v_x^{\text{max}}$ , while lateral and yaw-rate saturate at lower thresholds, consistent with our reward shaping for forward locomotion.}
  \vspace{-1em} 
  \label{fig:curr_rang_overview}
\end{figure}


\subsection{Experimental Setup (Policy Architecture)}

Teacher appears in two senses here: TransCurriculum Teacher selects task difficulty; while privileged teacher policy ($\pi_{T}$) selects actions, without sharing any parameters or gradient.


\textbf{Teacher policy.} Following \cite{margolis2024rapid}, \cite{rudin2022learning}, teacher policy ($\pi_{T}$) receives the current observation $o_t$ and the privilege environment parameters (for e.g., friction, restitution, base mass, joint friction, etc.) encoded by a learned embedding $e_\theta$  passed through a multilayer perceptron of size $[512,256,128]$ to produce joint commands.

\textbf{Student policy.} For deployment, the student policy ($\pi_{s}$) replaces $e_\theta$ with an implicit estimate from the history of the last $m$ time step $h_t = [x_{t-m}, x_{t-1}]$ performing system identification \cite{kumar2021rma}, \cite{lee2020learning} to mimic the teacher ($\pi_{T}$). 


\textbf{Reward Isolation.} We did not feed the reward history to either $\pi_{T}$ or $\pi_{S}$, it is used specifically by TransCurriculum module for bin selection. This ensures that performance gain is attributed to smart task scheduling at the curriculum level rather than additional policy input.

\textbf{Optimization.} Both teacher-student policies are optimized via  PPO \cite{schulman2017proximal}. TransCurriculum acts as meta-level sampler for curriculum for curriculum expansion.

\begin{figure}[h]
    \centering

    \begin{minipage}{0.42\linewidth}
        \centering
        \includegraphics[width=\linewidth]{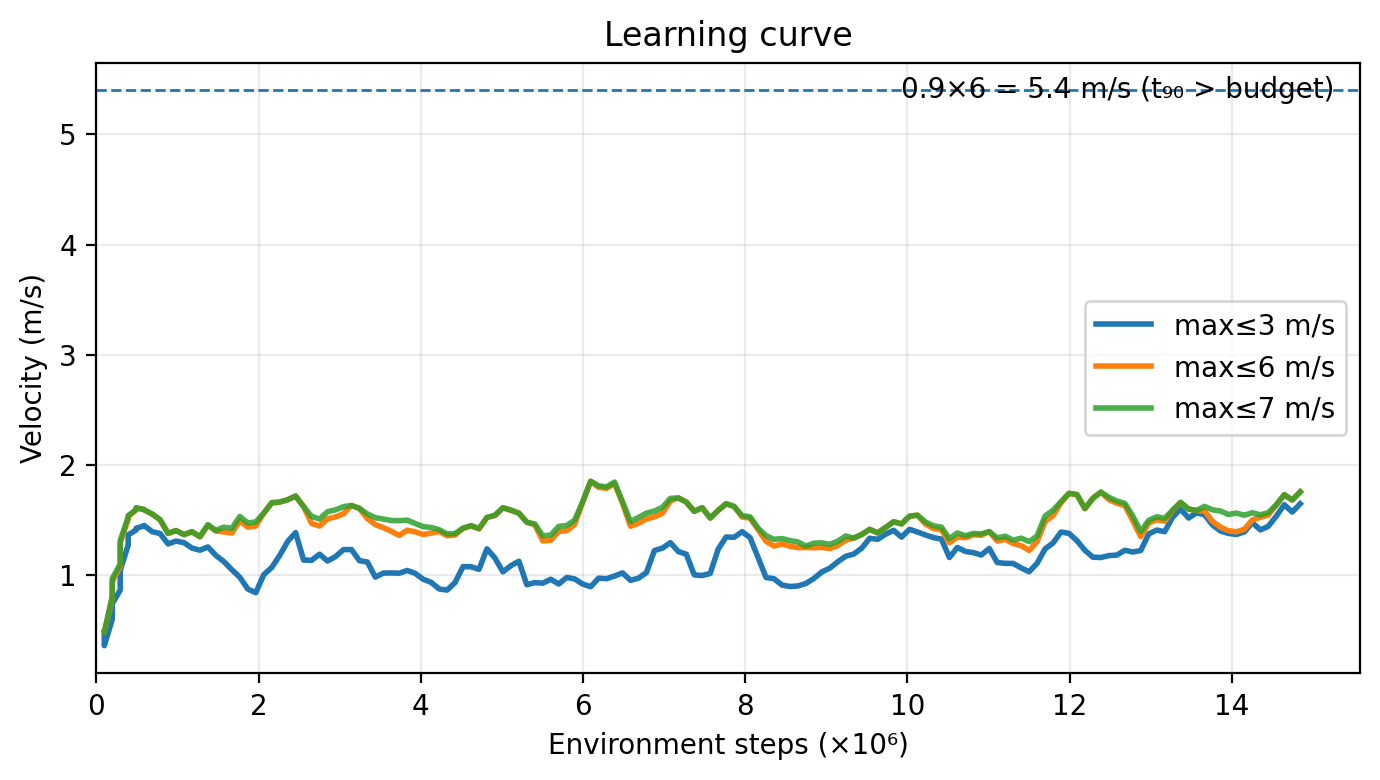}
    \end{minipage}
    \begin{minipage}{0.42\linewidth}
        \centering
        \includegraphics[width=\linewidth]{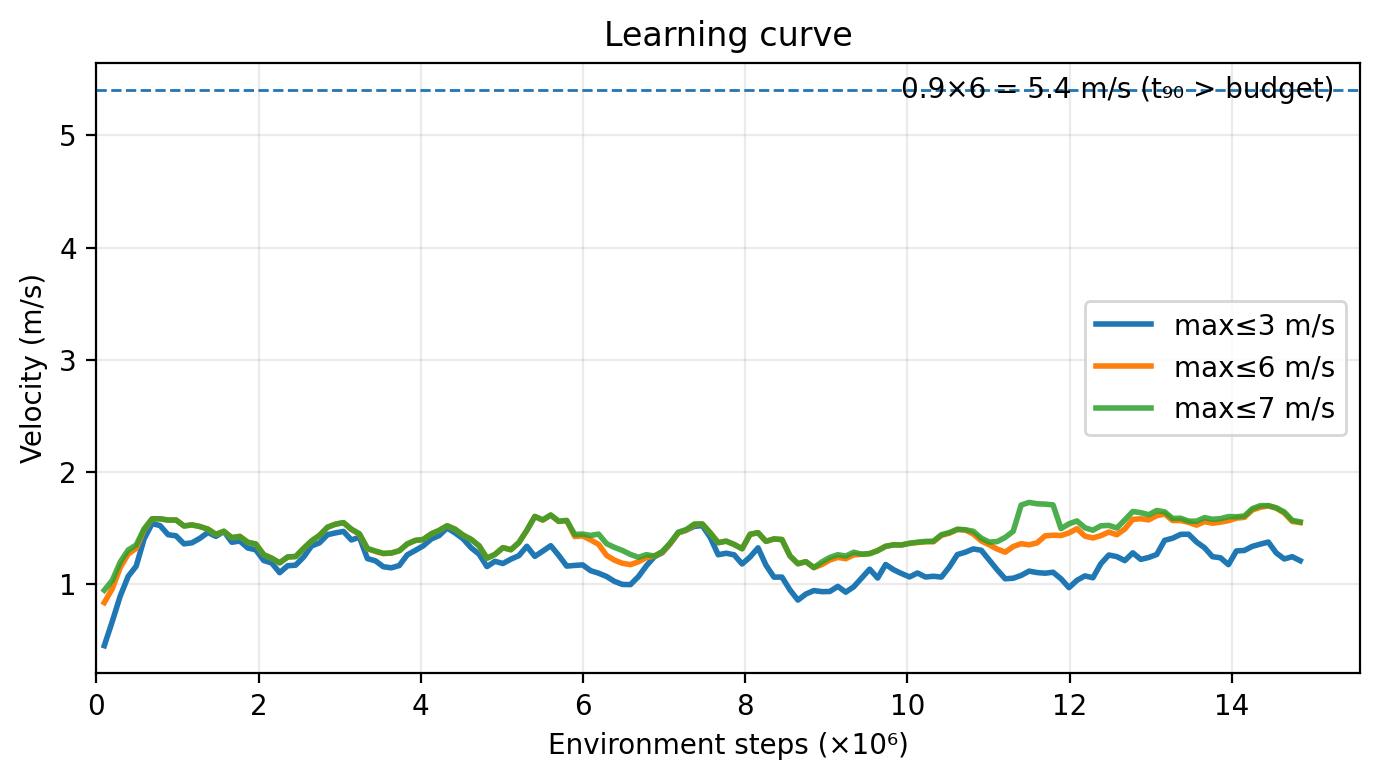}
    \end{minipage}
  
    \begin{minipage}{0.42\linewidth}
        \centering
        \includegraphics[width=\linewidth]{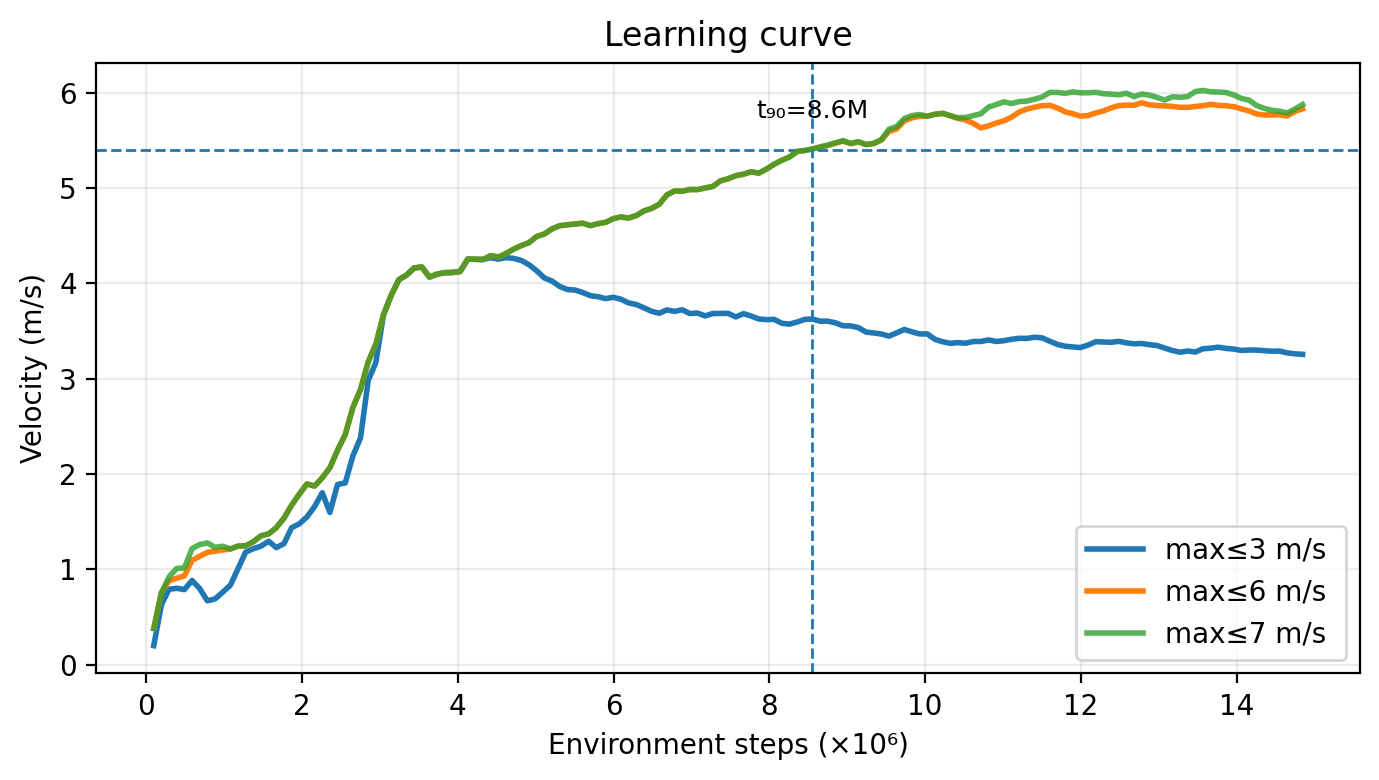}
    \end{minipage}
    \begin{minipage}{0.42\linewidth}
        \centering
        \includegraphics[width=\linewidth]{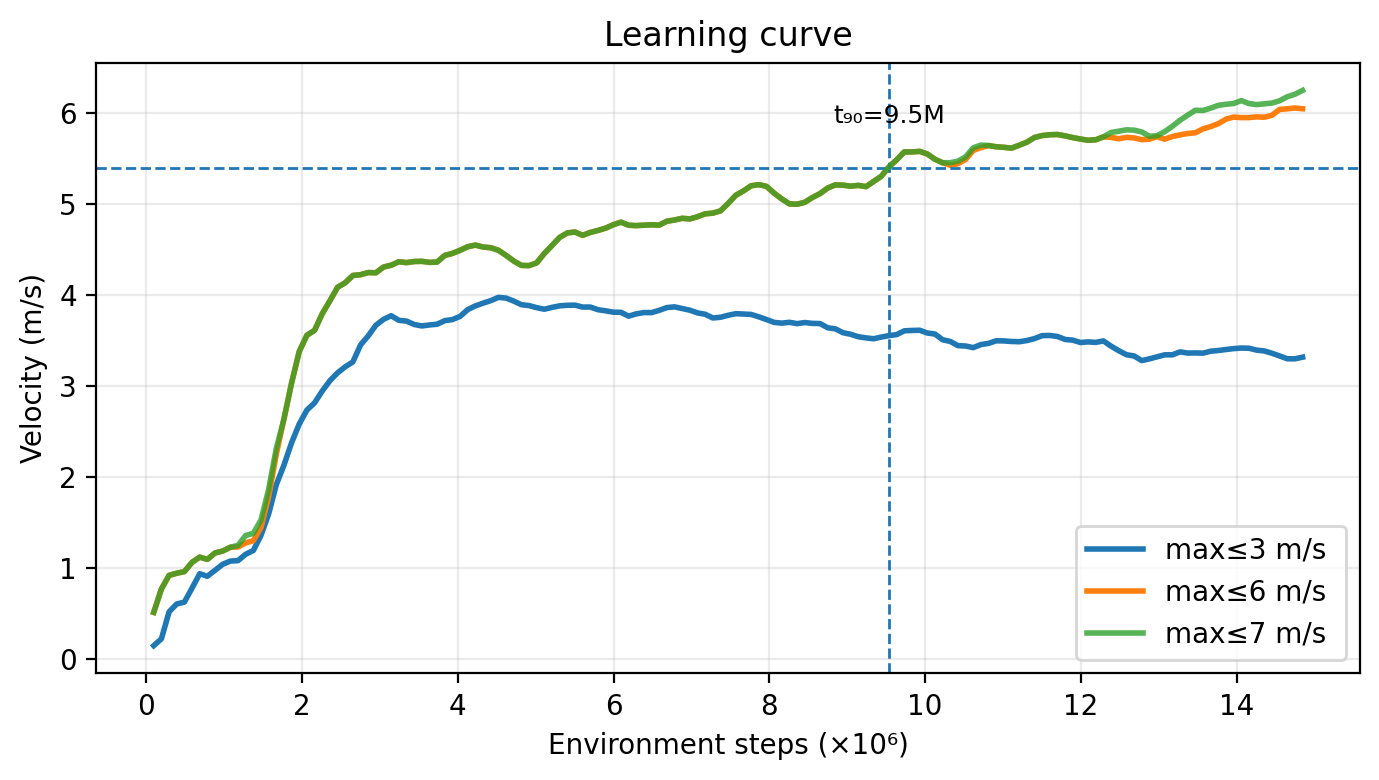}
    \end{minipage}           
   
    \caption{\textbf{Effect of curriculum bin selection criteria:} We train TransCurriculum for 250, 1000, 4000 and 6000 bins under identical training conditions and compare the learning curve (1500 PPO updates). Coarse binning (250/1000) do not reach the high-speed within the given training time, plateauing around $1.5$ -- $2.0$ m/s. The 4000-bin configuration achieves approximately 6 m/s and reaches 90\% of target speed within 8.6M environment steps, providing the best tradeoff between exploration and stable curriculum updates. With increasing resolution to 6000 bins increases velocity to $6-6.3$ m/s and the 90\% of target speed within 9.5M steps. We therefore use 4000 bins in our experiment unless noted otherwise.}
    \label{fig:hardware_robust}
\end{figure}

\begin{figure}[h]
    \centering

    \begin{minipage}{0.32\linewidth}
        \centering
        \includegraphics[width=\linewidth]{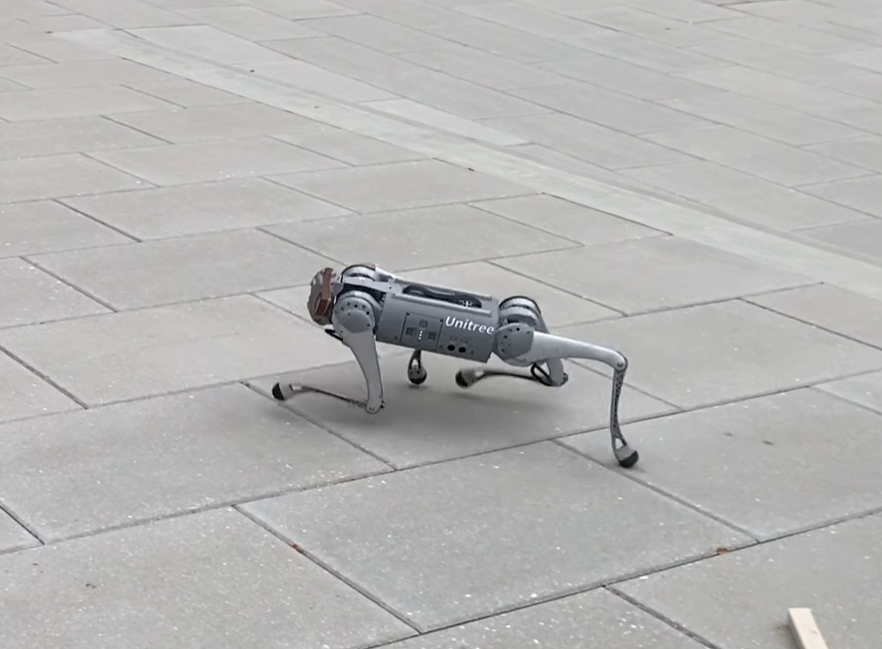}
    \end{minipage}
    \begin{minipage}{0.32\linewidth}
        \centering
        \includegraphics[width=\linewidth]{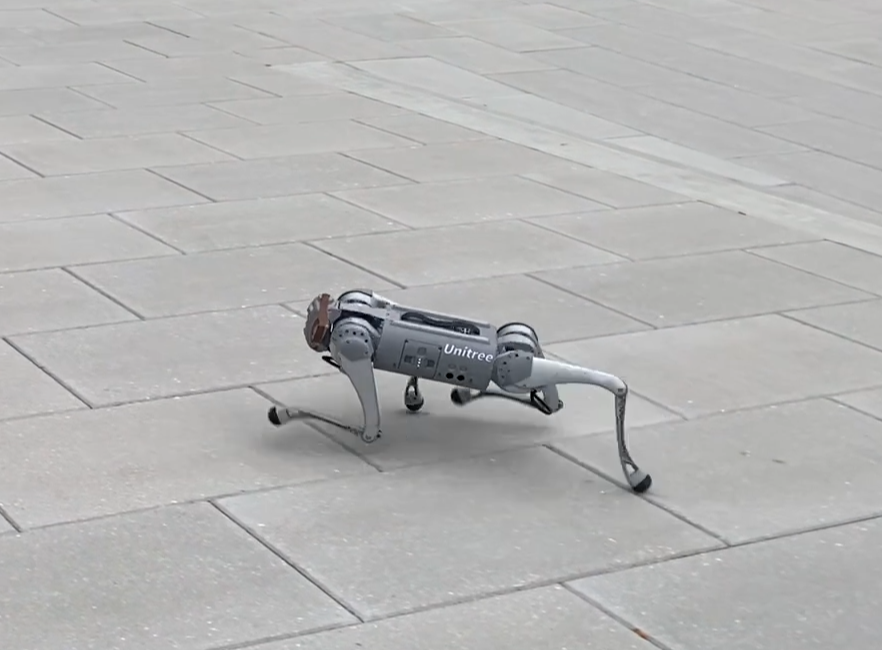}
    \end{minipage}
    \begin{minipage}{0.32\linewidth}
        \centering
        \includegraphics[width=\linewidth]{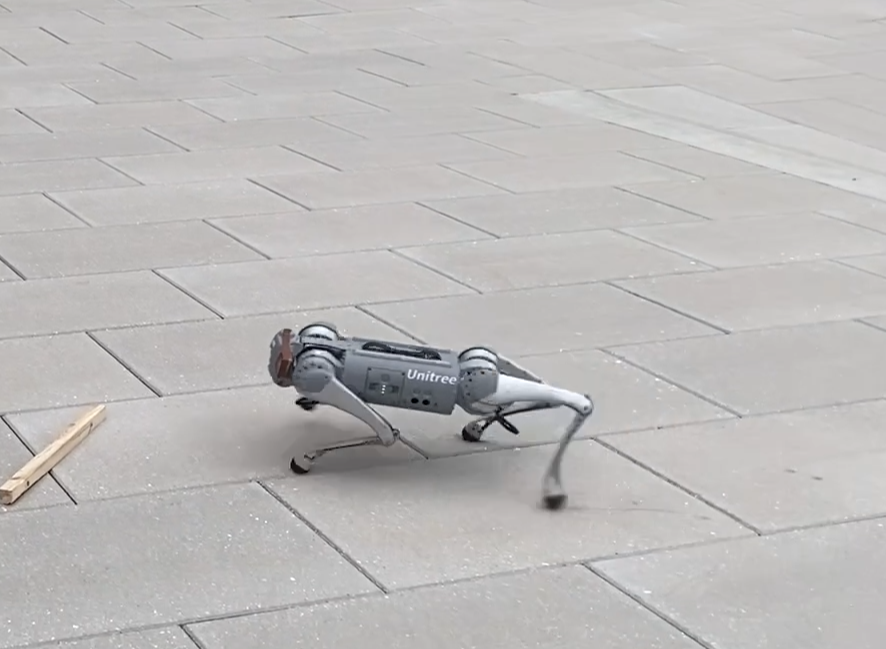}
    \end{minipage}
    




    \begin{minipage}{0.32\linewidth}
        \centering
        \includegraphics[width=\linewidth]{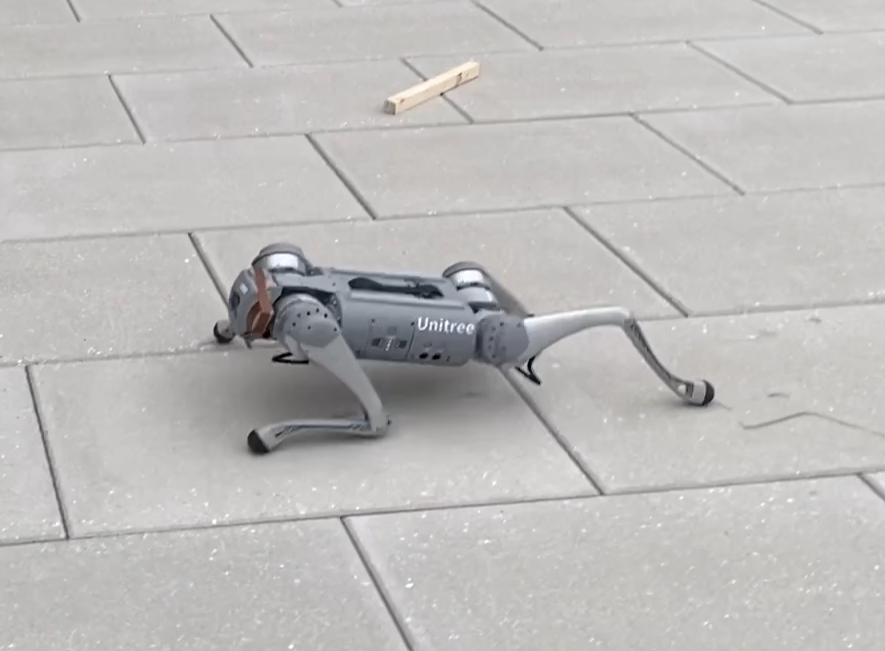}
    \end{minipage}
    \begin{minipage}{0.32\linewidth}
        \centering
        \includegraphics[width=\linewidth]{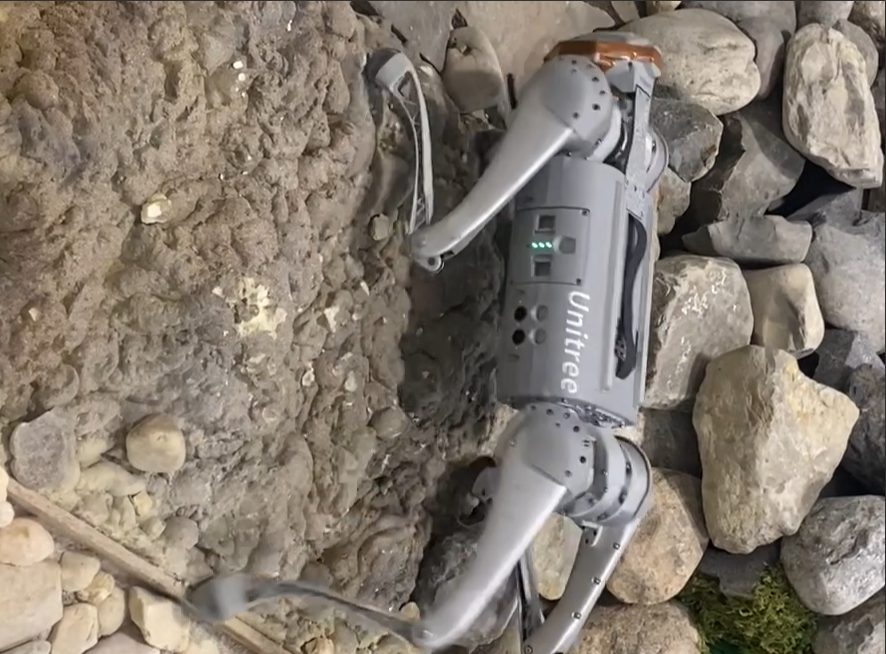}
    \end{minipage}
    \begin{minipage}{0.32\linewidth}
        \centering
        \includegraphics[width=\linewidth]{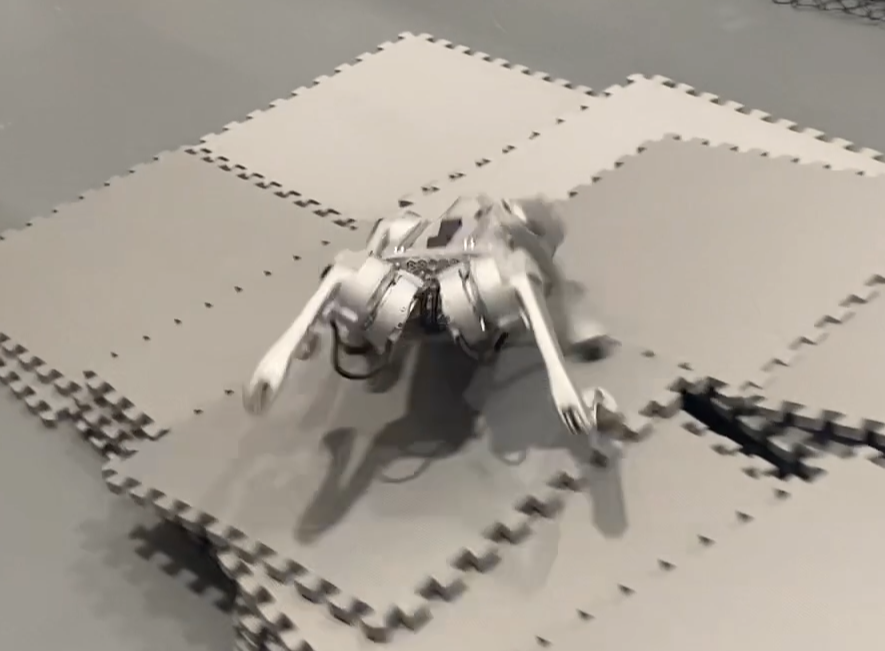}
    \end{minipage}
          
    \caption{\textbf{Disturbance recovery cases on diverse terrains (zero-shot Go1):} 
     The above cases demonstrates that our policy experiences a bump/trip or deviate laterally, and re-stabilizes to its normal gait within few seconds. Above examples demonstrate that TransCurriculum-trained policy remains functional without any additional finetuning under disturbances or terrain irregularities (See accompanying video).}
     \vspace{-1em} 
    \label{fig:hardware_robust}
\end{figure}

\subsection{Evaluation Metric}

\textbf{Cost of Transport (CoT)} \cite{schperberg2025energy}. The ratio of energy or power consumed by each joint to distance traveled \eqref{eq:cot}: 


\begin{equation}
\mathrm{CoT} = \frac{\int_0^T \sum_{j=1}^{N} \tau_j(t) \, \dot{q}_j(t) \, dt}{mg \, \Delta s}.
\label{eq:cot}
\end{equation}

Where $\tau_j(i)$ and $\dot{q}_j(i)$ are the torque and angular velocity of the joint $j$ at the time step $i$, and $mg \, \Delta s$ is the weight times distance traveled. Lower CoT indicates efficient locomotion.

\textbf{Stability score (S).} It is given by \eqref{eq:stability}:

\begin{equation}
S \;=\; 
\frac{1}{N}
\sum_{i=1}^{N} 
\sum_{k=1}^{K} 
W_k\, r_k(i),
\label{eq:stability}
\end{equation}

where $r_k(i)$ captures orientation, base height, angular velocity, linear velocity, joint position / velocity limit, self-collision, and torque limit constraints at each time step, with fixed weights $W_k$. Higher $S$ indicates greater stability.


\textbf{Task success rate.} The percentage of successful runs completed without a crash, tip, fall, or gait failure. We report the mean over 10 trials (real world) and 5 trials (simulation) with 95\% confidence interval.




\subsection{Baselines}

\textbf{Curriculum baselines:} We compare our TransCurriculum method against the major curriculum strategies used for locomotion: (i) Rapid-Motor Adaptation (RMA) \cite{kumar2021rma}: no explicit curriculum just gradually increases penalties on fixed schedule; (ii) RLvRL \cite{margolis2024rapid}: bin selection based on fixed rules for tracking velocity commands; (iii) Solo12 \cite{aractingi2023controlling}, \& Risky terrains\cite{zhang2024learning}: terrain difficulty (iv) CHRL \cite{li2024learning}:  hindsight based curriculum over command and terrain (v) LP-ACRL \cite{li2026scaling}: terrain curricula with learning progress signal. None of these approaches considers history modeling and Table II summarizes the curriculum signal, \& history-modeling for each method.






\textbf{History \& Non-history aware schedules.} To isolate the contribution of history modeling, we have compared both history (Transformer, RNN) and non history feedforward networks (MLP) as curriculum scheduler under identical training conditions like bin size, learning rate, PPO parameters, and reward functions. (Tables III). 



\section{Results \& Discussion}

\begin{figure*}[!t]
    \centering

    \begin{minipage}{\textwidth}
        \centering
        \includegraphics[width=0.19\textwidth]{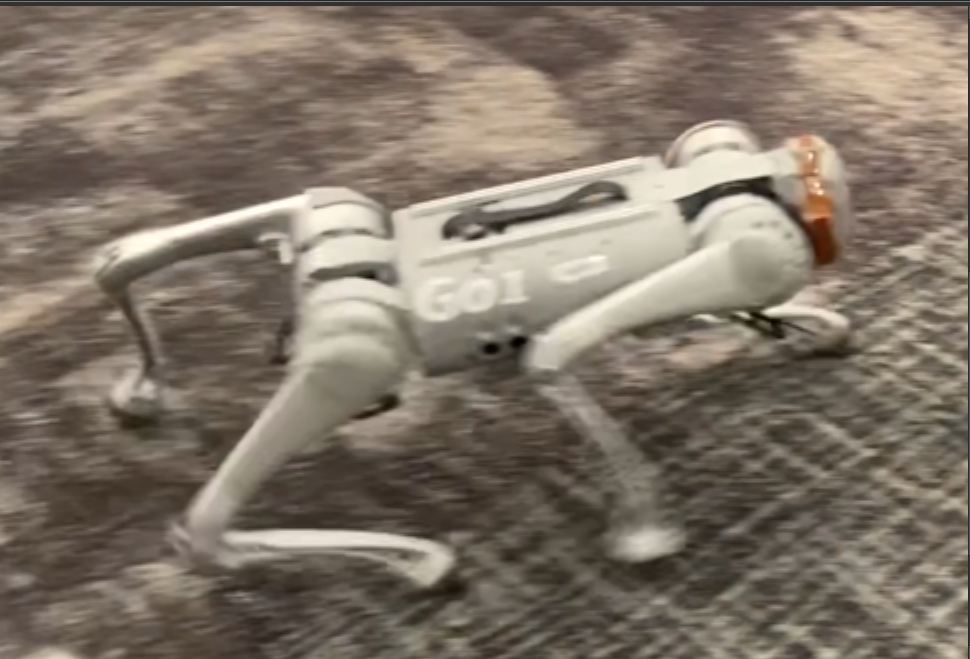}\hfill
        \includegraphics[width=0.19\textwidth]{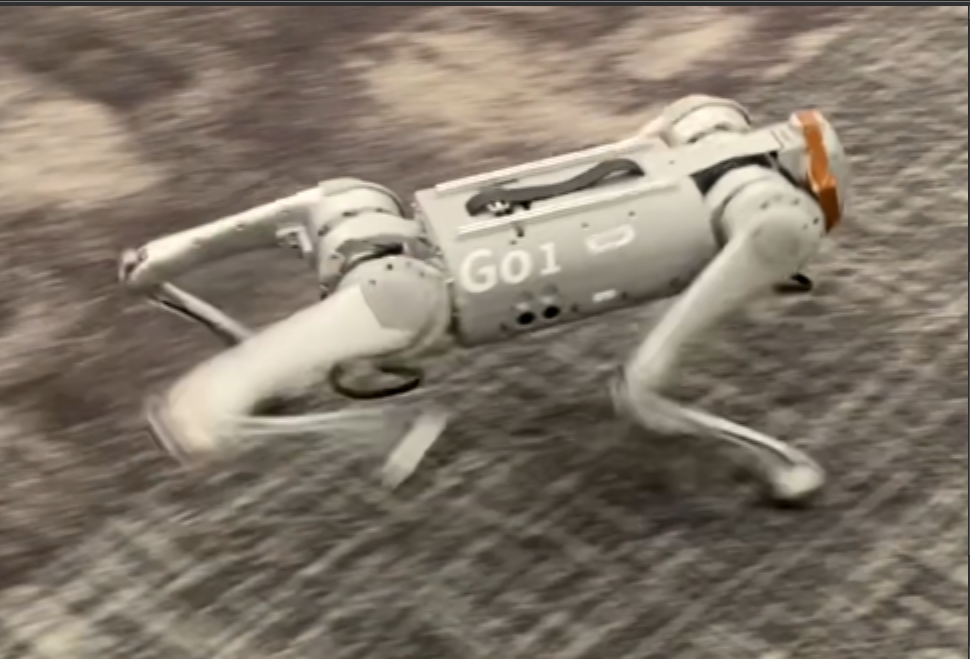}\hfill
        \includegraphics[width=0.19\textwidth]{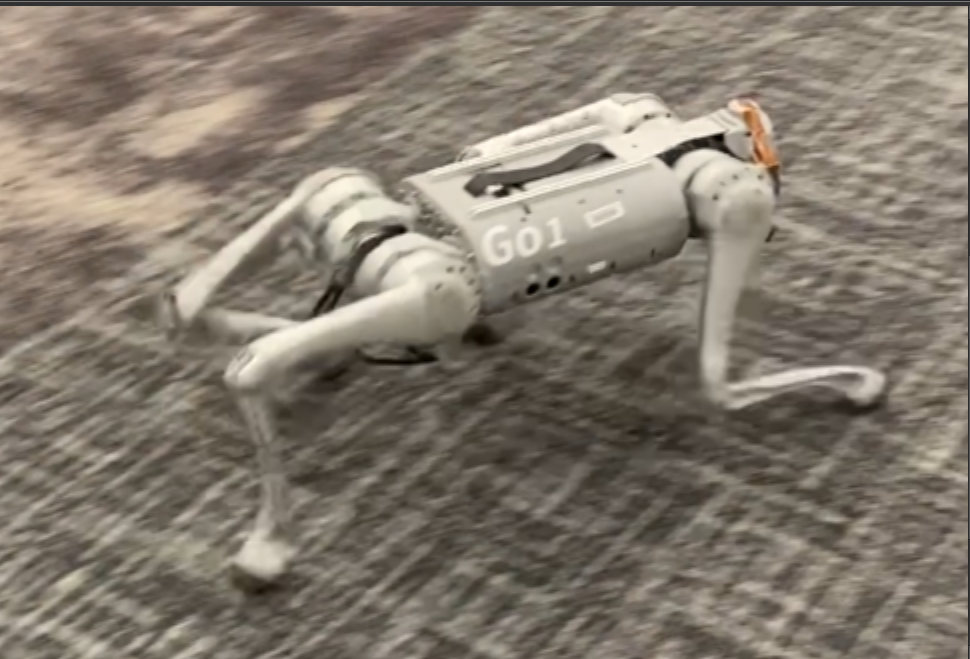}\hfill
        \includegraphics[width=0.19\textwidth]{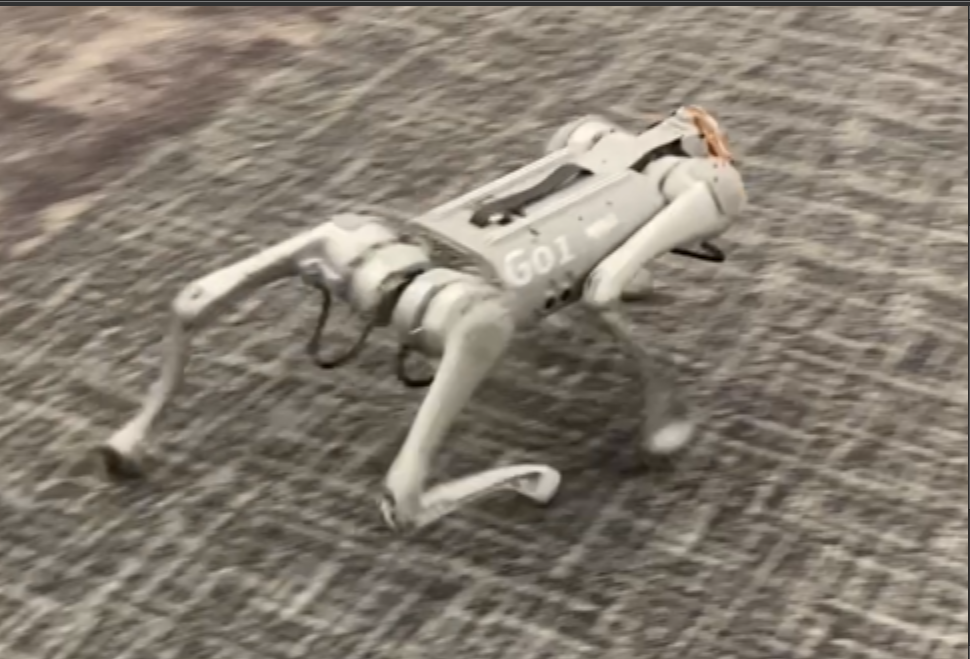}\hfill
        \includegraphics[width=0.19\textwidth]{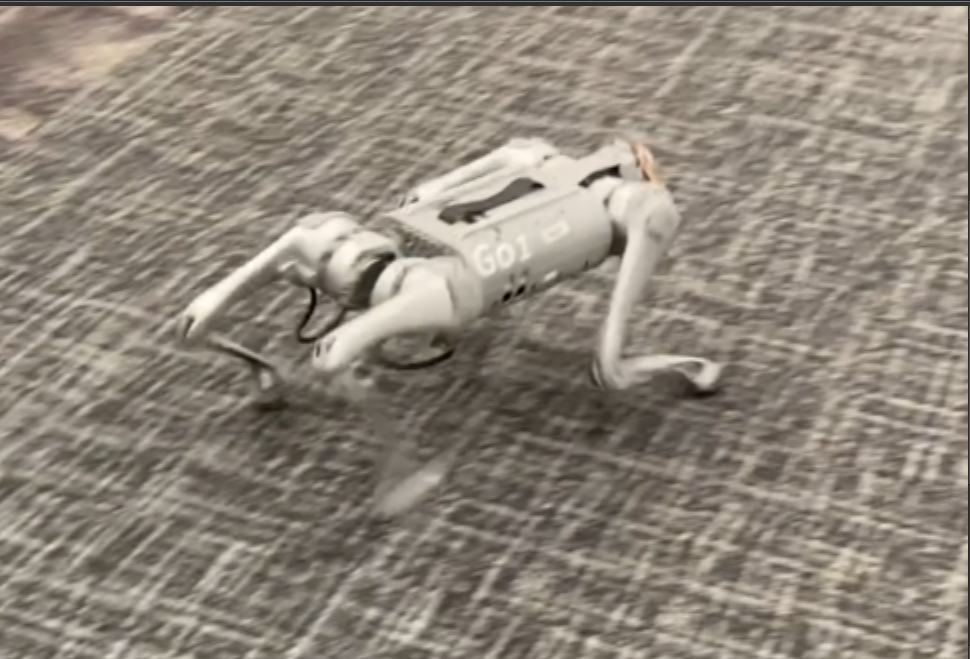}
    \end{minipage}

    \vspace{-0.05em} 
    \begin{minipage}{\textwidth}
        \centering
        \includegraphics[width=1.00\textwidth]{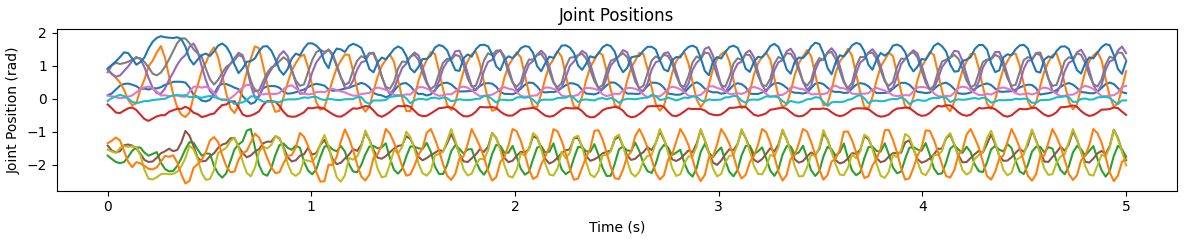}
    \end{minipage}


    \caption{\textbf{High-speed locomotion and stable gait with TransCurriculum.} 
    \textbf{Top:} zero-shot real-world deployment on Unitree Go1 on carpet, policy reaches $4.1 \pm 0.05$ m/s over a 3-8 m run with 90\% task success.
    \textbf{Bottom:} In simulation at $v^{cmd}_x = 7.0$, Go1 reaches $6.3 \pm 0.2$ m/s and the 12 joint-position time plot shows a consistent periodic pattern, indicating stable running gait at higher command velocity. Together these results substantiate that this is enabled by history-aware and multi-dimensional curriculum task sampling over commands, terrain and dynamics.}
    \vspace{-1em} 
    \label{fig:knit_fullwidth}
\end{figure*}


\subsection{Simulation Learning}

At $v^{cmd}_x = 7.0$ m/s, TransCurriculum reaches $6.3 \pm 0.2$ m / s and achieves the highest among curriculum baselines in our comparison table II. Compared to other command centric curriculum methods, like RLvRL (5.5 m / s) \cite{margolis2024rapid}, this corresponds to $+0.8~\mathrm{m/s}\ (\sim 14.55\%)$ improvement in maximum simulation speed.




\subsection{Zero-shot Hardware Transfer (Unitree Go1)}
We deploy our TransCurriculum trained policy zero-shot on a Unitree Go1 EDU robot (12 actuated joints). The policy runs on-board using IMU, joint encoders, and foot force sensors and outputs 12 joint target commands. Given hardware safety constraints, we tested up to $4.1 \pm 0.05$ m / s with a task success rate of 90\%. This exceeds the previous fastest reported RL-based zero-shot transfer of $3.9$ m / s on the MIT mini cheetah \cite{margolis2024rapid}. While other methods like DreamWaq\cite{nahrendra2023dreamwaq} report nearly $3.0$ m/s on Unitree A1, while CHRL \cite{li2024learning} report $3.45$ on a custom robot, Transcurriculum outperforms these and other curriculum baselines. Our key curriculum relevant findings and comparison are detailed in Table II.



\subsection{Terrain Robustness}
To evaluate robustness beyond the training distribution, we test TransCurriculum on Go1 across five terrain categories: rigid, deformable, irregular terrains, and moderate slopes.

\textbf{Rigid-Indoor:} Carpet (reference): $4.1 \pm 0.2$ m / s with a success rate 90\%. Tile: $3.1 \pm 0.4$ m/s, 100\% success; occasional foot slips reduce speed by $\approx 24.0\%$ but does not cause any failure.

\textbf{Rigid-Outdoor:} Cement: $3.3 \pm 0.3$ m / s with a success rate of 80\% (20.0\% below carpet).

\textbf{Deformable:} Grass: $1.8 \pm 0.2$ m / s, 70\% success (speed 56.0\% below the carpet).

\textbf{Irregular:} On pebbles: $ \approx 2.1 \pm 0.3$ m/s and 60\% success, with modest drift $0.3 \pm 0.1$. Broken rocks: $ \approx 1.5 \pm 0.4$ m / s and drift $0.3 \pm 0.2$, 50\% success (speed 48. 8\% and 63. 4\% below carpet).

\textbf{Slopes:} Cement ($15^\circ$): $2.7 \pm 0.3$ (90\%); Wood ($20^\circ$): $3.1 \pm 0.4$ m / s (80\%) (see Figures 1) (speed 34. 1\% and 24. 0\% below carpet).

Performance degrades with terrain compliance and irregularity, resulting in reduced speed, lateral deviation, and occasional failures. However, the system remains functional across all surfaces without any fine-tuning, indicating that multidimensional curriculum produces robust locomotion behavior. 


\subsection{Ablation Studies}

\textbf{History ablation (Table III):} We compare the history-aware curriculum (Transformer, RNN) against the non-history curriculum (MLP) under identical conditions (Table III). The stark difference is for $v^{cmd}_x = 7.0$ m/s, where a history-aware curriculum reaches $6.1$ -- $6.3$ m/s compared to $0.5$ m/s for non-history ones and fails the task (0\% success). The transformer provides the best trade-off in our runs, improving speed ($6.3$ vs $6.1$), stability ($2000$ vs $1800$) and task success (90\% vs 80\%) against the RNN baseline. We attribute the failure of non-history scheduler (MLP) to their inability to model temporal learning dynamics in the bin space. Without memory, they cannot identify bin that are useful from non-useful bins, which leads to poor sampling and progression.


\begin{table}[h]
\caption{{Non-history vs. History methods (\(n=5\) runs per method)}}
\centering

\resizebox{\columnwidth}{!}{%
\begin{tabular}{lcccc}
\hline
\textbf{Metric} & \textbf{Transformer} & \textbf{RNN} & \textbf{MLP} \\ \hline


$v_x$ m/s ($v^{cmd}_x = 7$) & 6.3 \textsuperscript{†} & 6.1 & 0.5 \\

$\omega_z$ rad/s & 1.25 & 1.28 & 0 \\
History-Aware & Yes & Yes & No \\
CoT & 2.60 & 5.20 & 100 x RNN \\

Stability (S) & 2000 & 1800 & 1100 \\
Task Success Rate & 90\% & 80\% & 0\% \\ \hline
\end{tabular}%
}
\textsuperscript{†}\,Reported for command only\\
\label{tab:history_vs_non}
\end{table}

\begin{table}[h]
\caption{Curriculum dimensionality and transfer loss ablation}
\centering
\resizebox{\columnwidth}{!}{%
\begin{tabular}{lcccc}
\hline
\textbf{Metric} & \textbf{Cmd Only}  & \textbf{Cmd + DR} & \textbf{Full (ours)} \\ \hline

Max sim velocity (m/s) & \textbf{6.3} & 6 & 5.8 \\
Stability score $(S)$ & 1850  & 1900 & \textbf{2000} \\
Task success rate (\%) & 70\% & 70\% & \textbf{90}\% \\
Sim-to-real \textsuperscript{†} (m/s) & 3.65  & 3.85 & \textbf{4.1} \\
Transfer loss  & 27\%  & 23\% & \textbf{18\%} \\ \hline
\end{tabular}%
}
\textsuperscript{†}\,Reported sim-to-real velocities for command velocity of 5 m/s.\\
\vspace{-1em} 
\label{tab:curriculum_ablation}
\end{table}

\textbf{Multidimensionality ablation (Table IV):} Table IV isolates the effect of multidimensionality on transfer by comparing command only, command + DR, and full joint space (command + DR + terrain), i.e. TransCurriculum (Full curriculum). Although the command only curriculum achieves fastest simulated speed ($6.3$ m/s), the full curriculum improves stability (1850 to 2000), increases the task success rate (70\% to 90\%) and reduces transfer loss (27\% to 18\%, nearly 33\% reduction). Our findings suggest that multidimensional primarily improves stability and transfer., while history-aware enables high speed training. Most prior work \cite{margolis2024rapid}, \cite{zhang2024learning}, \cite{li2024learning} treats velocity, terrain, and domain parameters as independent axis, but in the real-world these dimensions interact and training on a single axis leaves gaps, which show up as sim-to-real transfer loss.


\subsection{Failure Modes and discussion}

We identify three recurring failure patterns: (i) lateral deviation on a straight run of 2--8 m; (ii) slips and micro-stumbles on low friction and irregular terrains; and (iii) transient gait irregularities where the robot switches from trot to crawl (0.5--1.0 s) due to contact timing error.


These failures likely arise from (i) imperfect reward shaping focused for heading/straight line tracking, (ii) substrate mismatch, e.g., deformable/loose surfaces (like pebbles, tile, broken rocks) that were not modeled in the simulator, reducing traction and causing slips, and (iii) contact inference failure causing short gait dropout. Despite this, the policy recovers quickly from bumps and slips and maintains functional locomotion across all terrains tested. Addressing these modes through improved terrain modeling, reward shaping, and latent-aware control is a promising future work.

\section{Conclusion, Limitations, and Future Work}

We presented \textbf{TransCurriculum}, a multidimensional, history-aware transformer-based curriculum that enables fast and stable quadrupedal locomotion. We demonstrate the benefits of prioritizing curriculum expansion over joint space of velocity command, domain parameters, and terrain difficulty that it improves stability and reduces sim-to-real transfer loss (from 27\% for command only to 18\% for Transcurriculum (full)). In simulation, our approach achieves $6.3$ m/s and in zero-shot transfer on Go1 robot it reaches $4.1 \pm 0.05$ across multiple real-world terrains.


Our evaluations focus our policy on the quadrupedals, and we have not explored a single policy that is transferred across robot morphologies \cite{huang2020one}, and we do not evaluate bipedal/humanoid locomotion. For future work, the following directions remain promising i) bipedal/humanoid locomotion, ii) morphology generalization, and iii) curriculum to overcome uncertainty or contact related failures.














\balance

\bibliographystyle{IEEEtran}
\bibliography{refs}

\end{document}